\documentclass[10pt,twocolumn,letterpaper]{article}

\usepackage{iccv}      

\definecolor{iccvblue}{rgb}{0.21,0.49,0.74}
\usepackage[pagebackref,breaklinks,colorlinks,allcolors=iccvblue]{hyperref}

\usepackage{chngcntr}
\usepackage{graphicx}
\usepackage{multirow}
\usepackage{xcolor}
\usepackage{colortbl}
\usepackage{makecell} 
\usepackage{amssymb}
\usepackage{bbding}
\usepackage{pifont}
\usepackage[inkscapelatex=false]{svg}
\usepackage{wasysym}
\usepackage{utfsym}
\usepackage{algorithm}
\usepackage{algorithmic}
\usepackage{fontawesome}
\definecolor{my_green}{RGB}{51,102,0}
\definecolor{my_red}{RGB}{204, 0, 0}

\definecolor{ModelGreen}{RGB}{213,232,212}

\def\paperID{12256} 
\def\confName{ICCV}
\def\confYear{2025}

\title{Video-XL-Pro:\\ Reconstructive Token Compression for Extremely Long Video Understanding}

\author{
    Xiangrui Liu$^{1,2\dagger}$, 
    Yan Shu$^{3\dagger}$,
    Zheng Liu$^{2\dagger}$,
    Ao Li$^{1}$,
    Yang Tian$^{1}$,
    Bo Zhao$^{1*}$\\
    \small$^1$School of AI, Shanghai Jiao Tong University~~~
    \small$^2$Beijing Academy of Artificial Intelligence~~~
    \small$^3$University of Trento\\
    \small{\url{https://github.com/VectorSpaceLab/Video-XL/tree/main/Video-XL-Pro}}
}

\begin{document}
\maketitle
\begin{abstract}
Despite advanced token compression techniques, existing multimodal large language models (MLLMs) still struggle with hour-long video understanding. In this work, we propose Video-XL-Pro, an efficient method for extremely long video understanding, built upon Reconstructive Compression of Tokens (ReCoT), a learnable module that leverages self-supervised learning to generate comprehensive and compact video tokens. ReCoT introduces two key components: (i) Dynamic Token Synthesizer (DTS): DTS generates pseudo-video tokens from static image tokens by learning intra-token relationships, which are then used in masked video modeling. (ii) Semantic-Guided Masking (SGM): SGM adaptively masks redundant visual tokens to facilitate more effective reconstructive learning. To improve training efficiency in MLLMs fine-tuning, we introduce a video-specific dataset pruning strategy and design a simple yet Query-aware Selector that enables the model to precisely locate query-relevant video tokens.  
With only 3B parameters, Video-XL-Pro outperforms most 7B models trained on larger datasets across multiple long video understanding benchmarks. Moreover, it can process over 8K frames on a single A100 GPU while maintaining high-quality performance.

\end{abstract}


{
\renewcommand{\thefootnote}%
{\fnsymbol{footnote}}
\footnotetext[0]{$\dagger$ Equal contribution. * Corresponding authors.} 
}

\begin{figure}[h]
    \centering
    \includegraphics[width=0.45\textwidth]{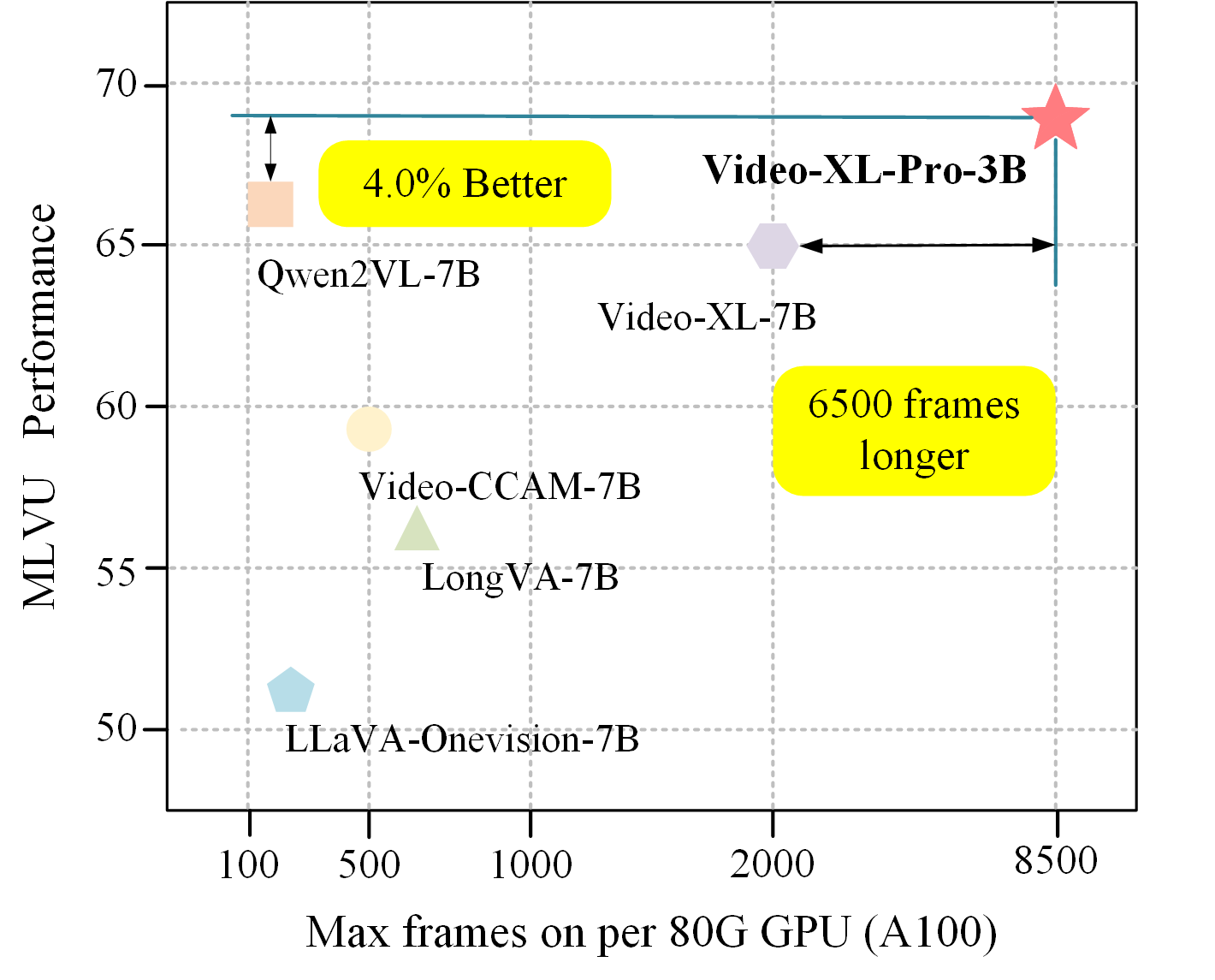}
    \caption{Compared with SOTA video understanding MLLMs, Video-XL-Pro achieves better accuracy and greater efficiency simultaneously.}
    \vspace{-5pt}
    \label{fig:radar}
\end{figure}

\section{Introduction}
\label{sec:intro}

Recently, large language models (LLMs)~\cite{ChatGPT,openai2023gpt4,touvron2023llama} have demonstrated remarkable capabilities in text understanding. Through visual instruction tuning~\cite{liu2023visual_llava,videollava}, which aligns visual and linguistic modalities, multimodal large language models (MLLMs) have achieved promising results in visual information reasoning. More recently, there has been a growing effort to adapt MLLMs for video understanding tasks, including video captioning and question answering~\cite{li2023videochat,llamavid,videollama}.


However, existing methods face significant challenges when processing long videos (exceeding three minutes). Compared to single images, longer videos require sampling more frames, leading to an increased number of visual tokens, which in turn results in higher memory costs and exceeds the inherent limitations of LLMs. To address this issue, various compression modules have been proposed to reduce token counts generated by the visual encoder~\cite{maaz2023videochatgpt,llamavid,weng2024longvlm,malmm2024}. While these methods facilitate longer video understanding, several concerns remain. First, the quality of compressed tokens remains agnostic, as there is no established criterion to evaluate their effectiveness directly. Second, existing methods typically employ a fixed compression ratio (e.g., 8 × or 16 ×), disregarding the information intensity gap between different videos.

One potential solution to this challenge is self-supervised representation learning, where Masked Autoencoders (MAE)~\cite{he2021maskedautoencodersscalablevision} have demonstrated remarkable success by enabling a substantial portion of an image to be masked and subsequently reconstructed with high fidelity.  This concept has been extended to video modeling (e.g. VideoMAE~\cite{tong2022videomaemaskedautoencodersdataefficient}), leveraging temporal redundancy to enable efficient video reconstruction. Inspired by this, we ask: \emph{Can we develop a learnable and adaptive video compression method that enables reconstructive token compression?} Ideally, if such compact video tokens can preserve the original content sufficiently for accurate reconstruction, they would greatly benefit MLLMs by enhancing both efficiency and effectiveness in video understanding. Based on this insight, we propose Video-XL-Pro, which proposes the reconstructive token compression (ReCoT) method. Specifically, ReCoT comprises two key components.

$\bullet$  Dynamic Token Synthesizer (DTS). Integrating VideoMAE with video MLLMs is a nontrivial challenge. Existing video MLLMs typically rely on image encoders such as CLIP~\cite{radford2021learningtransferablevisualmodels} or SigLIP~\cite{zhai2023sigmoidlosslanguageimage} to extract video embeddings. However, these models struggle to capture dynamic motion over time, particularly when processing high-FPS videos. While video encoders are trained to capture global video features, they are inherently limited in handling arbitrary frames, restricting their adaptability to diverse video structures. To address this limitation, we propose the Dynamic Token Synthesizer (DTS), which incorporates lightweight spatio-temporal attention blocks to enhance intra-token relationships. DTS actively learns dynamic motions within videos, enabling more effective representation learning.

$\bullet$  Semantic-guided Masking (SGM). Performing self-supervised reconstructive learning in the video domain typically requires large-scale data, leading to time-consuming training. Moreover, previous studies have primarily relied on random-based masking, which often select redundant or low-information regions, such as backgrounds, hindering effective video representation learning. To address these limitations, we propose Semantic-Guided Masking (SGM), which adaptively masks tokens by considering both intra-frame information density and inter-frame spatio-temporal significance. Instead of selecting tokens based on naive similarity calculations, SGM identifies and preserves structurally and semantically meaningful regions through a learnable mechanism, facilitating a more efficient and effective reconstruction process.


By pre-training on joint image and video data, ReCoT generates comprehensive and compact tokens, facilitating MLLM fine-tuning. To mitigate the computational and time costs associated with fine-tuning for video understanding, we propose a video-specific dataset pruning strategy that selects high-quality video data based on frame-wise similarity differences. By filtering out redundant or low-informative videos, our method ensures that the model is trained on diverse, representative, and information-rich video samples, leading to improved fine-tuning efficiency and better generalization for MLLMs. Moreover, video understanding requires precise retrieval of useful information for a given query. To address this, we design a simple yet effective Query-aware selector based on retrieval-augmented generation (RAG), enabling the model to locate query-relevant video tokens for improved understanding.

The effectiveness of Video-XL-Pro is validated in the following aspects. First, despite being trained on less data, the 3B Video-XL-Pro model outperforms most 7B video MLLMs across several long video understanding benchmarks, including MLVU~\cite{zhou2024mlvu}, Video-MME~\cite{videomme}, and LongVideoBench~\cite{wu2024longvideobench}. Second, it achieves an optimal balance between effectiveness and efficiency, processing over 8K frames on a single A100 GPU while attaining more than 95\% accuracy in the Needle-in-a-Haystack evaluation. Third, the proposed ReCoT achieves comparable or superior performance to the strong SigLIP baseline on mainstream image and video benchmarks, including ImageNet~\cite{russakovsky2015imagenet}, UCF101~\cite{soomro2012ucf101}, and Kinetics~\cite{carreira2018quovadisactionrecognition}.

\section{Related work}
\label{sec:formatting}

\paragraph{Multimodal Large Language Models.}
Inspired by the remarkable success of Large Language Models (LLMs), recent Multimodal Large Language Models (MLLMs) have integrated visual encoders to extract image embeddings, which are then adapted to match the LLM's token dimension through specialized connector modules~\cite{huang2023languageneedaligningperception,alayrac2022flamingovisuallanguagemodel,li2023blip2,zhu2023minigpt4,liu2023visual_llava}. The pioneering work Kosmos-1~\cite{huang2023languageneedaligningperception} establishes an end-to-end method that seamlessly integrates visual inputs with LLMs through a unified training paradigm. Subsequent approaches, such as Flamingo~\cite{alayrac2022flamingovisuallanguagemodel} and BLIP-2~\cite{li2023blip2}, utilize cross-attention layers and Q-Former modules, respectively, to combine visual and textual representations. More recent innovations, exemplified by MiniGPT-4~\cite{zhu2023minigpt4} and LLaVA~\cite{liu2023visual_llava}, further streamline the integration process by directly projecting visual features into the LLM's embedding space via MLPs, offering more efficient paradigms for MLLMs construction.

\begin{figure*}[t]
    \centering
    \includegraphics[width=\textwidth]{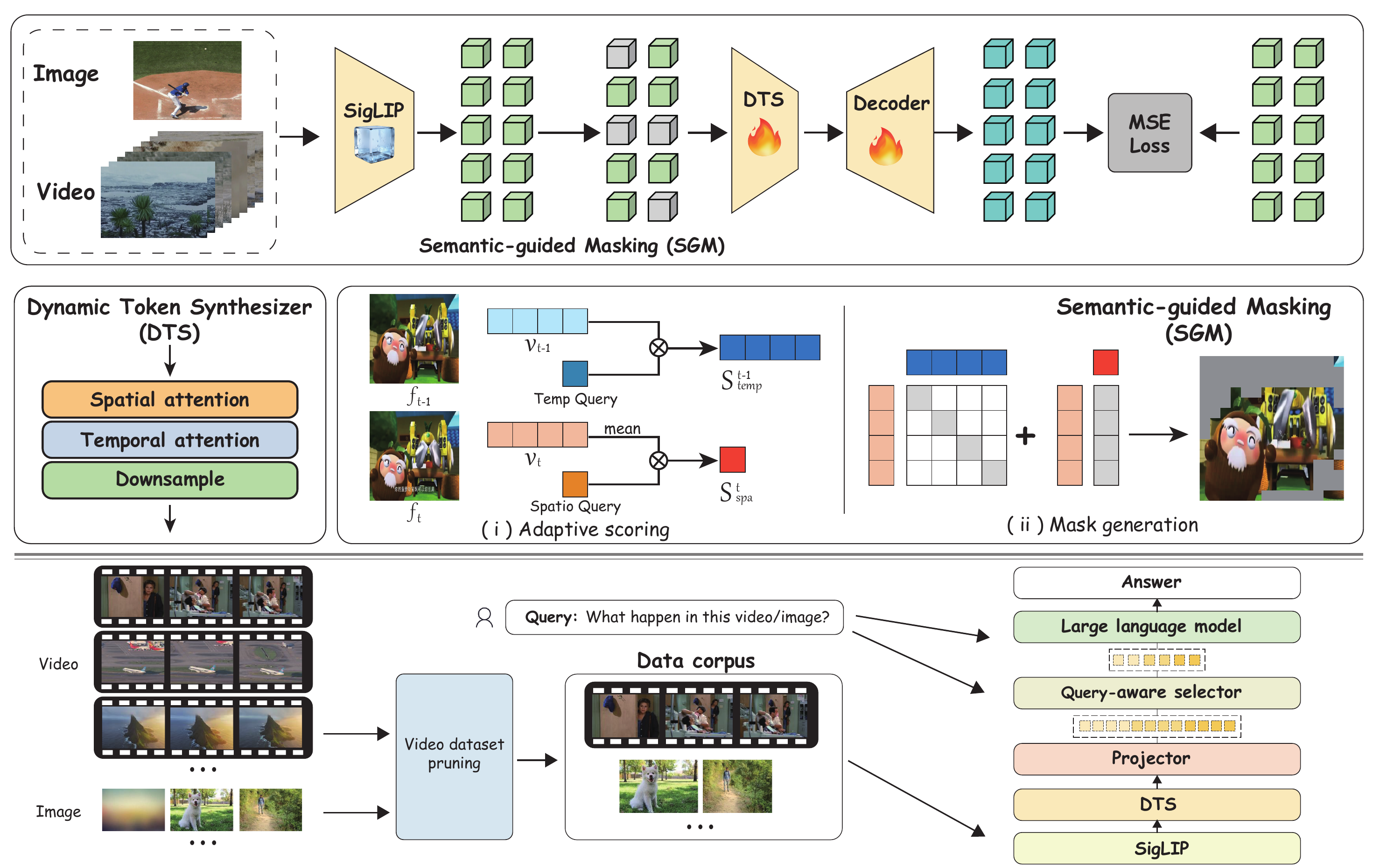}
    \caption{Overview of Video-XL-Pro. The top part is the reconstructive token compression, in which we propose reconstructive token compression (ReCoT) to generate comprehensive and compact tokens. The bottom part is the MLLM training stage, where we propose the video dataset pruning strategy and query-aware selector to improve efficiency.}
    \label{fig:framework}
    \vspace{-0.3cm}
\end{figure*}

\paragraph{Video MLLMs.}
Many studies have extended the capabilities of MLLMs to video understanding by encoding multi-frame features and concatenating them for uniform interpretation~\cite{moviechat2023,he2024malmmmemoryaugmentedlargemultimodal,llamavid,chatunivi,weng2024longvlm,shu2024videoxl}. To enhance temporal relationships between frame tokens, a temporal attention mechanism, such as Q-Former~\cite{li2023blip2}, is applied to merge informative tokens. To address long video understanding, various compression-based methods have been proposed to reduce visual tokens before they reach the LLM. Leveraging offline encoding and online decoding, MovieChat~\cite{moviechat2023} and MA-LMM~\cite{he2024malmmmemoryaugmentedlargemultimodal} introduce a memory mechanism to store key frames and perform streamlined decoding based on the given query. LLaMA-VID~\cite{llamavid} utilizes an additional text decoder to embed textual queries for cross-attention among frame features while compressing context tokens into a single token per frame. Chat-UniVi~\cite{chatunivi} and LongVLM~\cite{weng2024longvlm} aim to merge spatial and temporal redundant information, facilitating efficient long video understanding. Unlike visual-centric compression methods, Video-XL~\cite{shu2024videoxl} leverages the KV sparsification capacity in LLMs, introducing a visual summarization token to compress video segment tokens into compact representations.

\section{Method}

\subsection{Overview}
The overview of Video-XL-Pro is divided into two stages: self-supervised pertaining for reconstructive token compression and MLLM training, as illustrated in Figure \ref{fig:framework}. In the first stage, our goal is to develop a robust visual encoder capable of generating compact and comprehensive visual tokens using self-supervised learning methods. In Section \ref{sec:pretrain}, we provide a detailed explanation of the two key components: Dynamic Token Synthesizer (DTS) and Semantic-Guided Masking (SGM). In the second stage, we utilize pretrained SigLIP and DTS as our visual encoder and follow VLM for fine-tuning. Additionally, we introduce a video dataset pruning strategy to enhance training efficiency, along with a Query-Aware Selector to expand the maximum input context length supported by VIdeo-XL-Pro, as detailed in Section \ref{sec:MLLMtraining}.

\subsection{Reconstructive Token Compression}
\label{sec:pretrain}
During ReCoT training, we freeze the parameters of SigLIP. The input images are first processed by SigLIP to generate tokens, which are then passed through the DTS module. Under the guidance of SGM, the tokens output by SigLIP are compressed by a factor of four. Subsequently, we use a decoder to reconstruct the compressed features back into tokens. During this process, we employ Mean Squared Error (MSE) as the reconstruction loss supervision. For training data, we utilize 250K images from Bunny-695K \cite{he2024bunny} and 750K long videos from LLaVA-Video \cite{li2024llavaonevision}.

\textbf{Dynamic Token Synthesize (DTS).} The DTS module is positioned after SigLIP, serving two key functions: (i) summarizing redundant tokens and (ii) capturing dynamic motions from videos. Specifically, DTS is constructed with spatio-temporal attention blocks and 3D convolutional layers, enabling it to merge tokens output by SigLIP from four frames into a single frame's worth of tokens along the temporal dimension. We utilize frozen SigLIP output tokens and employ an imbalanced encoder-decoder architecture, where the decoder contains half the parameters of the encoder, to reconstruct the tokens output by SigLIP. With our proposed Semantic-Guided Masking (SGM) strategy (discussed in Section X), we efficiently train the DTS module in just 48 A100 GPU hours.


\textbf{Semantic-Guided Masking (SGM).} Appropriately masking the input is a common training technique in Autoencoder (AE) models, as it enhances the model's generalization ability. Masking during training forces the model to focus on the overall structure of the data, thereby learning more representative features. However, as the proportion of masking increases, the required training data grows exponentially, making the training cost prohibitively high. On the other hand, using a very small proportion of masking or no masking at all can hinder the model's ability to learn global features. To address this, we propose a Semantic-Guided Masking (SGM) strategy, which adaptively masks redundant features in videos and images, thereby guiding the DTS to perform efficient reconstruction training. 
Specifically, the semantic-guided mask involves two sets of learnable tokens: Temp Query and Spatio Query. First, Temp Query is multiplied with the tokens of the previous frame to obtain $V_{t-1}$, while Spatio Query is multiplied with the semantic average of the current frame's tokens to obtain $V_{t}$. Subsequently, $V_{t-1}$ and $V_{t}$ are used to compute attention with the tokens of the current frame. Finally, the two sets of attention scores are summed to obtain the final token scores. As shown in Algorithm 1, under the guidance of the token scores, we randomly mask low-scoring tokens during training, thereby directing the model to focus on key regions in the video. With the guidance of SGM, we can efficiently train DTS using only a small amount of data.

\begin{algorithm}[h]
\caption{Semantic-Guided Masking} \label{al}
\begin{algorithmic}[1]
\STATE $\textbf{Input}$: Siglip tokens
\STATE $\textbf{Output}$: All tokens score of Semantic-Guided Masking
\STATE $videofeature \gets Siglip(video)$
\STATE $V_{t-1} \gets torch.cat((zero,videofeature[:-1]))$
\STATE $V_{t} \gets Mean(videofeature,dim=1) $
\STATE $S_{temp} \gets V_{t-1} \otimes Temp Query$
\STATE $S_{spa} \gets V_{t} \otimes Spatiio Query$
\STATE $score_{temp} \gets \textbf{torch.einsum}(videofeature, S_{temp})$
\STATE $score_{spa} \gets videofeature \otimes S_{spa}$
\STATE $tokenscore=score_{temp}+score_{spa} $
\end{algorithmic}
\end{algorithm}

\begin{table*}[t]\small
\centering
\vspace{-4mm}
\addtolength\tabcolsep{-2.4pt} 
\resizebox{1.0\linewidth}{!}{
\begin{tabular}{lcc|c|c|cc|c|c|ccc}
\toprule
\multicolumn{1}{c}{\multirow{2}{*}{Model}} & \multicolumn{1}{c}{\multirow{2}{*}{Size}} & \multicolumn{1}{c|}{\multirow{2}{*}{Training Data (M)}} & \multicolumn{1}{c|}{\multirow{2}{*}{MLVU Dev}}  & \multicolumn{1}{c|}{\multirow{2}{*}{MLVU Test}} & \multicolumn{2}{c|}{VideoMME} & \multicolumn{1}{c|}{\multirow{2}{*}{VNBench}}  & \multicolumn{1}{c|}{\multirow{2}{*}{LongVideo}} & \multicolumn{1}{c}{\multirow{2}{*}{TempCompass}} \\

\multicolumn{1}{c}{} & \multicolumn{1}{c}{} & \multicolumn{1}{c|}{} & \multicolumn{1}{c|}{}  & \multicolumn{1}{c|}{}  & W/o sub & W sub & \multicolumn{1}{c|}{} & \multicolumn{1}{c|}{} & \multicolumn{1}{c}{} & \multicolumn{1}{c}{} & \multicolumn{1}{c}{} \\ \midrule

\rowcolor{gray!15}\multicolumn{12}{c}{\textbf{Proprietary Models}} \\
GPT-4V~\cite{openai2023gpt4} & - & - & 49.2  &43.3  &59.5& 63.3 & 48.9   & 59.1 & -   \\
GPT-4o~\cite{gpt4o} & - & - & \textbf{64.6} & \textbf{54.9}  &71.9& 71.2 & 64.4  & \textbf{66.7}  & \textbf{73.8} \\
Gemini-1.5-Pro~\cite{reid2024gemini} & - & -&- &  -  &\textbf{75.0}& \textbf{81.3} &  \textbf{66.7}  & 64.0  & 67.1 \\
\midrule

\rowcolor{gray!15}\multicolumn{12}{c}{\textbf{Small Size MLLMs}} \\ 
InternVL2.5-2B~\cite{chen2025internvl2.5} & 2B & -& 61.4  & -&51.9 & 54.1 & - &52.0  &- \\
InternVL2.5-4B~\cite{chen2025internvl2.5} & 4B & -& \textbf{68.3}  & -&\textbf{62.3} & 63.6 & - &55.2  &- \\
VideoChat-Flash-2B~\cite{Li2025videochatflash} & 2B &- & 65.7  & - &57.0 & 63.9 & -  &\textbf{58.3}  & - \\
VideoLLaMA3-2B~\cite{zhang2025videollama3frontiermultimodal} & 2B & 15.6M/22M/19M/5.7M & 65.4  & - &59.6 & 63.4 & - &57.1  &63.4 \\
LongVU-3.2B~\cite{Shen2024longvu} & 3B & 3.2M/0.5M & 55.9  &-  & 51.5 & -  & - & -  & -  \\
Apollo-3B~\cite{Orr2024apollo} & 3B & 0.2M/0.4M/3.2M & 65.0 & -&58.4 & 60.6 & - & 55.1  & 62.5 \\
Qwen2.5-VL-3B~\cite{bai2025qwen25vltechnicalreport} & 3B & - & 68.2 & - & 61.5 & \textbf{67.6} & - & 54.2  & \textbf{64.4} \\

\rowcolor{gray!15}\multicolumn{12}{c}{\textbf{Open-source MLLMs}} \\ 
LongVA-7B~\cite{longva} & 7B & - & 56.3  &41.1 & 52.6 & 54.3 & 41.5 & 47.8 &  57.0  \\
Pixtral-12B~\cite{agrawal2024pixtral12b} & 7B & -   & - & -  & 40.7 & 47.5  & -& - & -\\
LongVU-7B~\cite{Shen2024longvu} & 7B & 3.2M/0.5M & 65.4 &-  & 60.6 & - & - & -  & -\\
Oryx1.5~\cite{liu2024oryx} & 7B & - & 63.8 & -  & 59.0 & - & -  & -& -  \\
LLaVA-OV-7B~\cite{li2024llavaonevision} & 7B & - & 64.7  &-  & 58.2 & 61.5 & - & -  & - \\
Video-XL~\cite{shu2024videoxl} & 7B & 2.0M/1.0M & 64.9  & 45.5   &55.5 &61.0  &61.6   &50.7 & - \\ 
TimeMaker-8B~\cite{Chen2024timemaker} & 8B & 45M & - & -  & 57.3 & 62.8 & -  & -& 60.4  \\
Apollo-7B~\cite{Orr2024apollo}& 7B & 0.2M/0.4M/3.2M & 67.3  &-   &\textbf{61.3} & 63.3& - & \textbf{58.5} & 64.9  \\
\midrule

\rowcolor{ModelGreen}\textbf{Video-XL-Pro} & 3B & \textbf{2.0M/1.0M} & \textbf{70.6} & \textbf{47.0}   &60.0 &\textbf{64.3}  &\textbf{63.0}  & 56.7 & \textbf{65.7} \\ 
\bottomrule
\end{tabular}}

\vspace{-2mm}
\caption{Experimental results on mainstream video benchmarks. ``LongVideo."  refer to LongVideoBench Bench, respectively. $\dag$ indicates that the results on VNBench and LongVideoBench were reproduced using their official weights.} 
\vspace{-2mm}
\label{tab:mlvu} 
\end{table*}

\begin{table}[h]\small
\centering
\resizebox{1.0\linewidth}{!}{
\begin{tabular}{lccccc}
\toprule
\multicolumn{1}{c}{\multirow{2}{*}{Method}} & \multicolumn{1}{c}{\multirow{2}{*}{Size}} & \multicolumn{4}{c}{\textbf{V-STaR}} \\
\cmidrule(lr){3-6}
 & \multicolumn{1}{c}{} & $R@1_{(\text{IOU}=0.3)}$ & $R@1_{(\text{IOU}=0.5)}$ & $R@1_{(\text{IOU}=0.7)}$ & mIoU\\
\midrule
VTimeLLM~\cite{huang2024vtimellm} & 7B & 25.24 & 10.88 & 3.15 & 17.13 \\
TimeChat~\cite{timechat} & 7B & 17.80 & 8.68 & 3.48 & 12.01\\
TRACE\cite{guo2025vtgllmintegratingtimestampknowledge} & 7B & 28.53  & 14.17 & 6.73 & 19.74\\
\midrule
InternVL-2.5~\cite{chen2025internvl2.5} & 8B & 11.98 & 4.87 & 2.34 & 8.72 \\
Qwen2.5-VL-7B~\cite{bai2025qwen25vltechnicalreport} & 7B & 17.03 & 8.92 & 3.72 & 11.48 \\
Video-LLaMA3~\cite{zhang2025videollama3frontiermultimodal} & 7B & 35.73 & 19.80 & 8.68 & 22.97 \\
\midrule
\rowcolor{ModelGreen}\textbf{Video-XL-Pro} & 3B & \textbf{40.72} & \textbf{25.78} & \textbf{15.58} & \textbf{25.07}\\
\bottomrule
\end{tabular}}
\caption{Results on temporal grounding task, Performance Comparison on V-STaR.}
\label{tab:time1}
\end{table}

\begin{table}[h]\small
\centering
\resizebox{1.0\linewidth}{!}{
\begin{tabular}{lccccc}
\toprule
\multicolumn{1}{c}{\multirow{2}{*}{Method}} & \multicolumn{1}{c}{\multirow{2}{*}{Size}} & \multicolumn{4}{c}{\textbf{Charades-STA}} \\
\cmidrule(lr){3-6}
 & \multicolumn{1}{c}{} & $R@1_{(\text{IOU}=0.3)}$ & $R@1_{(\text{IOU}=0.5)}$ & $R@1_{(\text{IOU}=0.7)}$ & mIoU\\
\midrule
VTimeLLM~\cite{huang2024vtimellm} & 7B & 51.0  & 27.5 & 11.4 & 31.2 \\
TimeChat~\cite{timechat} & 7B & - & 32.2 & 13.4 & - \\
TimeSuite~\cite{zeng2024timesuite} & 7B & 69.9 & 48.7 & 24.0 & - \\
\midrule
Qwen2.5-VL-7B~\cite{bai2025qwen25vltechnicalreport} & 7B & - & - & - & 43.6 \\
VideoChat-Flash~\cite{Li2025videochatflash} & 7B & 72.5 & 51.4 & 26.4 & 48.0 \\
\midrule
\rowcolor{ModelGreen}\textbf{Video-XL-Pro} & 3B & \textbf{80.3} & \textbf{64.1} & \textbf{42.7} & \textbf{51.8}\\
\bottomrule
\end{tabular}}
\caption{Results on temporal grounding task, Performance Comparison on Charades-STA.}
\label{tab:time2}
\end{table}

\subsection{MLLM Training}
\label{sec:MLLMtraining}
Following the success of the MLLM paradigm, which leverages visual instruction tuning to align vision and language representations, Video-XL-Pro consists of three main modules: the LLM backbone, Vision Encoder, and Cross-Modality Projector. We select Qwen-2.5 \cite{yang2024qwen2} with 3B parameters as our LLM, due to its strong language understanding and reasoning capabilities. 
For the Vision Encoder, we utilize the pretrained visual encoder, which consists of SigLIP and DTS, enabling the generation of comprehensive and compact visual tokens. Following LLaVA \cite{liu2023visual_llava}, we employ a two-layer MLP with a GELU activation function as a Cross-Modality Projector to align the Vision Encoder and LLM. Besides, we propose the Query-aware selector for selecting informative tokens that permanent to the given query, and propose video dataset pruning for improving efficiency.

\textbf{Query-aware Selector.} Although DTS can further compress tokens output by SigLIP, generating efficient and compact representations to reduce visual redundancy, the computational cost remains high when processing extremely long videos (typically exceeding 1,000 frames). To address this, we design a Query-Aware Selector based on the Retrieval-Augmented Generation (RAG) framework. Specifically, we encode the input text query using an LLM and pass it through a trainable MLP layer to compute attention scores with the visual tokens. Following the Query-Aware Selector strategy, we use these attention scores to guide the selection of visual tokens, ensuring that the input token length remains within the LLM's allowable limit while filtering out tokens irrelevant to the current query. Specifically, the Query-Aware Selector randomly drops 5\% to 30\% of the tokens. However, during inference, the selector is only activated when the input context length exceeds the LLM's limit.

\textbf{Video Dataset Pruning.} Compared to image-based question answering, annotated video data is scarce, particularly for long videos. Although many studies attempt to scale up video understanding datasets using synthetic methods, two major challenges remain unresolved. First, the quality of synthetic data is uncertain, as low-quality samples may hinder effective model training. Second, training with video data is computationally expensive and time-consuming, as it involves processing a larger number of relevant tokens during training. Therefore, we propose a Video Dataset Pruning Strategy. First, we divide all video data into distinct subsets based on their task type, including video captioning, multiple-choice question answering, and open-ended question answering. Next, for each video sample, we uniformly sample 20 frames and use SigLIP to compute intra-frame cosine similarity. Based on a predefined threshold, each subset is compressed by filtering out videos with high semantic redundancy. Finally, we retain high-quality videos with informative intensity to enhance training efficiency.

To further enhance our model's performance on long videos, we employ Variable Sampling during training. In previous training pipelines, videos are usually uniformed sampling which can not capture the key nuance between frames. Moreover, their short sampling length hinder the model generalization to longer videos. Thus, for video training data, we employ variable sampling (with a maximum of 360 frames) to enhance the model's generalization capability for long videos. For short videos, we randomly sample at 2 fps, 3 fps, etc., while for long videos, we use a fixed sampling rate of 1 fps. The Figure \ref{fig:data_durations} below illustrates the duration distribution of our training set videos after adopting variable sampling compared to fixed sampling.

\begin{figure}[h]
    \centering
    \includegraphics[width=0.37\textwidth]{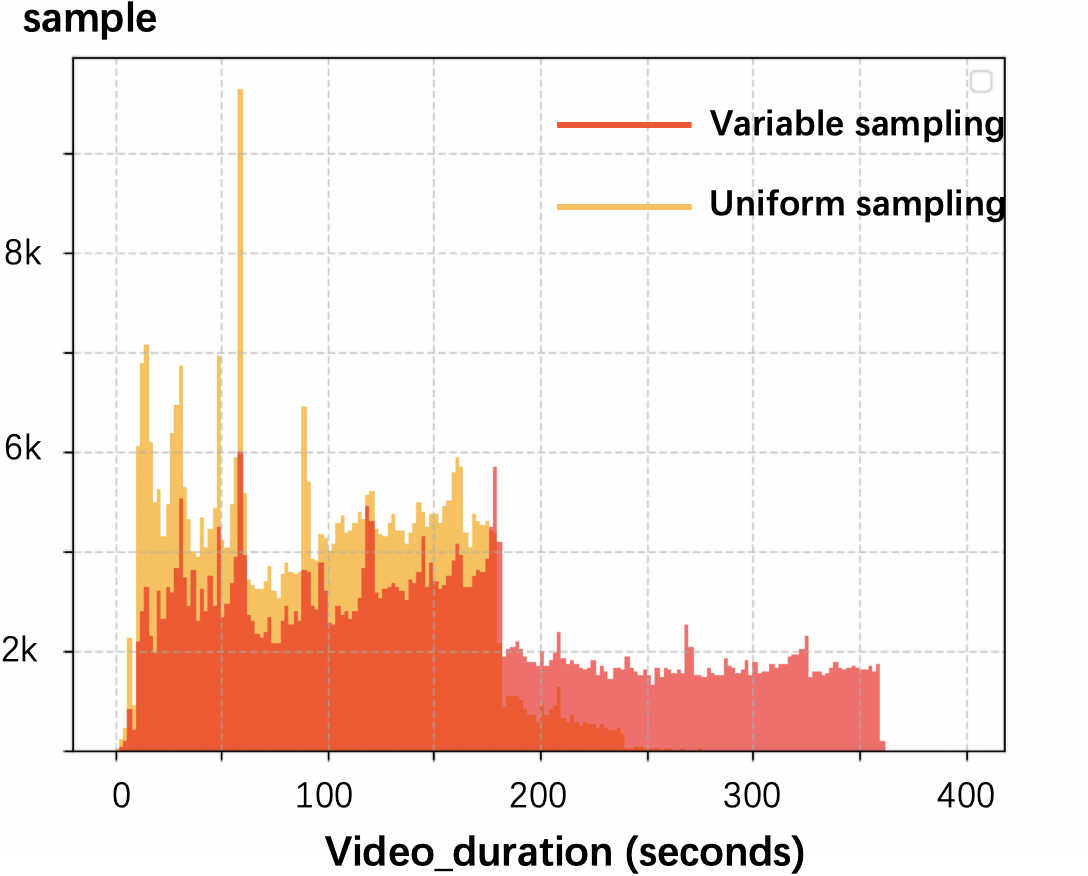}
    \caption{Training data distribution of uniform and variable sampling.}
    \vspace{-5pt}
    \label{fig:data_durations}
\end{figure}
\section{Experiment}

\subsection{Training Data}

We use large-scale image-text and video-text datasets for pre-training, aligning vision and language modalities. For image-text pairs, we collect 2M samples, removing low-quality data with noisy labels while retaining those with detailed captions. For video-text pairs, we apply video-length-based filtering, retaining shorter videos with minimal scene changes. Ultimately, we collect 2M image samples and 400K video samples from LLaVA-OneVision~\cite{li2024llavaonevision}, LAION-2M~\cite{he2024bunny}, OpenViD~\cite{nan2025openvid1m}, and Charades~\cite{sigurdsson2016charades}. 
In the fine-tuning stage, the following datasets are collected: Bunny~\cite{he2024bunny}, Sharegpt-4o~\cite{sharegpt4o}, and OCR-VQA~\cite{Anand2019ocrvqa}, These datasets are combined with our pruned video data, which contains NExT-QA~\cite{xiao2021next}, Sharegpt-4o~\cite{sharegpt4o}, CinePile~\cite{rawal2024cinepile}, LLaVA-178K~\cite{zhang2024videoinstructiontuningsynthetic}, VCGPlus~\cite{maaz2024vcgplus} for training.

\begin{figure*}[t]
    \centering
    \vspace{-0.3cm}
    \includegraphics[width=1.0\textwidth]{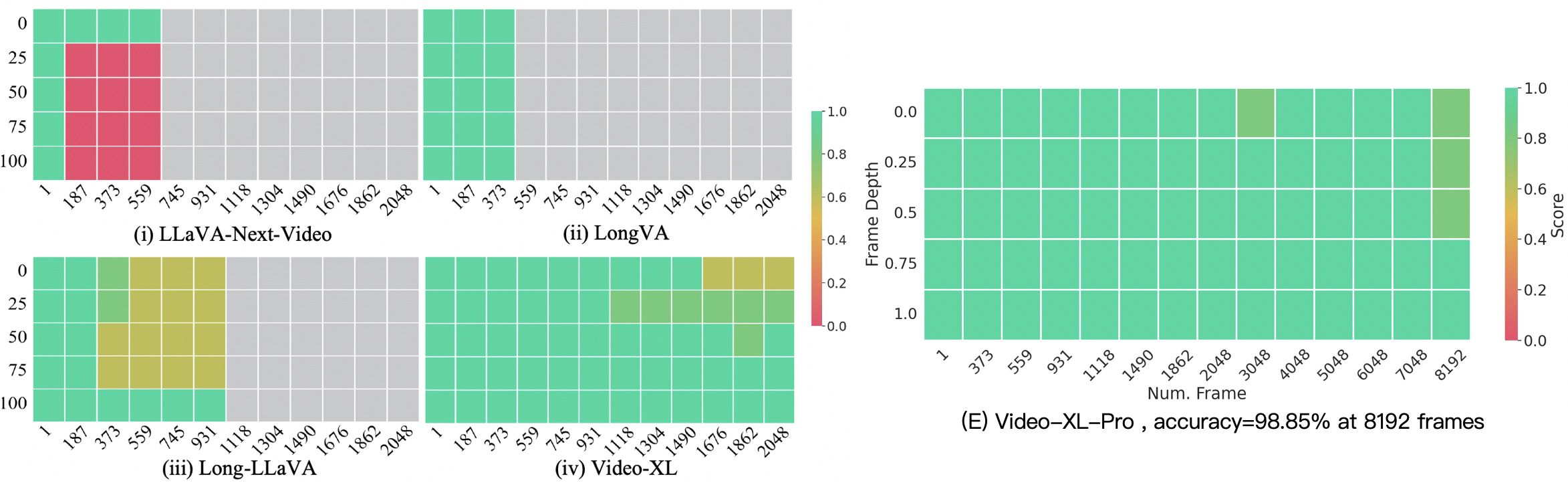}
    \caption{Results on the Needle-in-a-haystack evaluation within a single A100 80GB GPU. The x-axis represents the total number of frames in the video haystack. The y-axis shows the position where the needle image is located. Gray grids mean ``OOM'.  }
    \label{fig:needle}
    \vspace{-0.3cm}
\end{figure*}

\subsection{Implementation Details}
In the pretraining stage, only the projector is trained, with a learning rate of 5e-4. In the fine-tuning stage, all parameters are trained, with the learning rate set to 1e-5. The batch sizes for pretraining and fine-tuning are 8 and 1, respectively. We employ a cosine learning rate schedule with a warm-up ratio of 0.03 in both stages. In the fine-tuning stage, videos are sampled at 1 FPS, and if the video length exceeds 240 frames, we uniformly sample 240 frames. All experiments are conducted on $32\times$ A800-80GB GPUs.


\begin{figure}[h]
    \centering
    \includegraphics[width=0.5\textwidth]{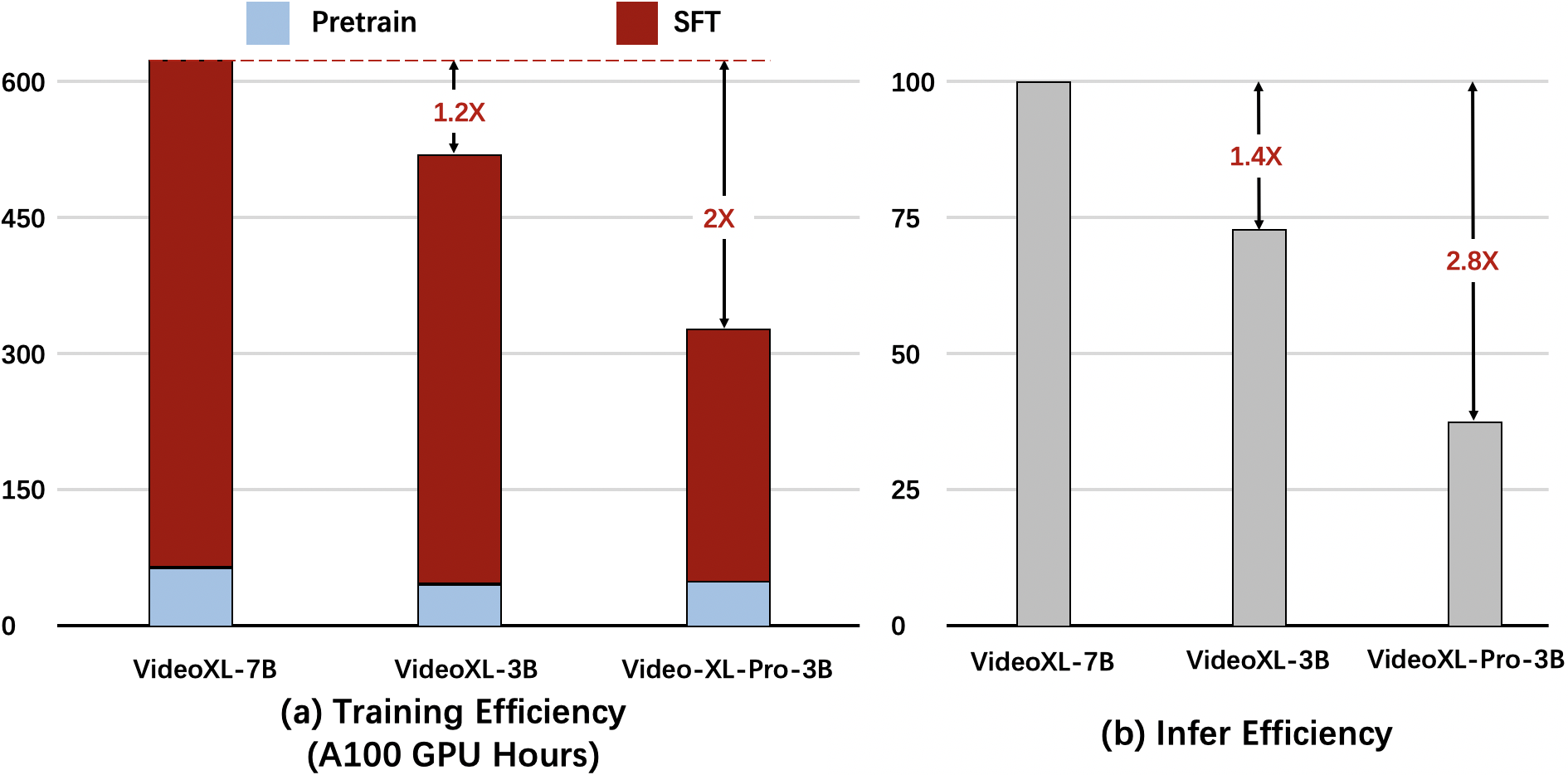}
    \caption{The training and inference efficiency of Video-XL-Pro.}
    \vspace{-8pt}
    \label{fig:data_duration}
\end{figure}

\subsection{{Benchmarks and Metrics}}
We empirically evaluate the effectiveness of Video-XL-Pro based on several popular long video understanding benchmarks. 1. MLVU~\cite{zhou2024mlvu}, a comprehensive benchmark that consists of multiple choice and generation tasks.
2. Video-MME~\cite{videomme}, another extensive benchmark covering videos of diverse genres and lengths (short, medium, and long).
3. VNBench~\cite{vnbench}, a synthetic benchmark focused on assessing models’ ability to handle video tasks, such as retrieval, ordering, and counting.
4. LongVideoBench~\cite{wu2024longvideobench}, a benchmark designed for tasks that require precise retrieval and reasoning over detailed multi-modal information within extended inputs.
5. Tempcompass~\cite{liu2024tempcompass}, a benchmark designed for evaluating visual-language models' temporal understanding of long videos.
We further evaluate the model’s timestamp awareness using the Charades-STA~\cite{huang2024vtimellmcharadessta} temporal grounding dataset and V-STaR~\cite{cheng2025vstarbenchmarkingvideollmsvideo} long video temporal grounding dataset.
\subsection{Main Results}

We present the performance of Video-XL-Pro on popular long video benchmarks in Table \ref{tab:mlvu}. Notably, it outperforms existing methods on both the Dev and Test set of MLVU. Despite having only 3B parameters, its performance on MLVU even surpasses GPT-4o, as well as well-known open-source models with similar parameter scales, such as Qwen2.5-VL-3B and InternVL2.5-4B. It also outperforms the majority of existing 7B models, including Apollo 7B, which was trained on 3.2M sft data, by 3.3\%, demonstrating highly competitive results.
For Video-MME, Video-XL-Pro achieves accuracies of 60.0\% and 64.3\% under the “no subtitles” and “with subtitles” settings, respectively, reaching state-of-the-art levels compared to advanced models. In TempCompass, Video-XL-Pro achieves the best performance among open-source models, even outperforming Qwen2.5-VL-3B by 1.3\% and Apollo 7B by 0.8\%, especially considering it is trained with far less data. It also achieves competitive results on LongVideoBench. Last but not least, although Video-XL-Pro is primarily designed for long video understanding tasks, it also excels in short video tasks, achieving competitive results on the VNBench benchmark.


\begin{figure*}[t]
    \centering
\includegraphics[width=1\textwidth]{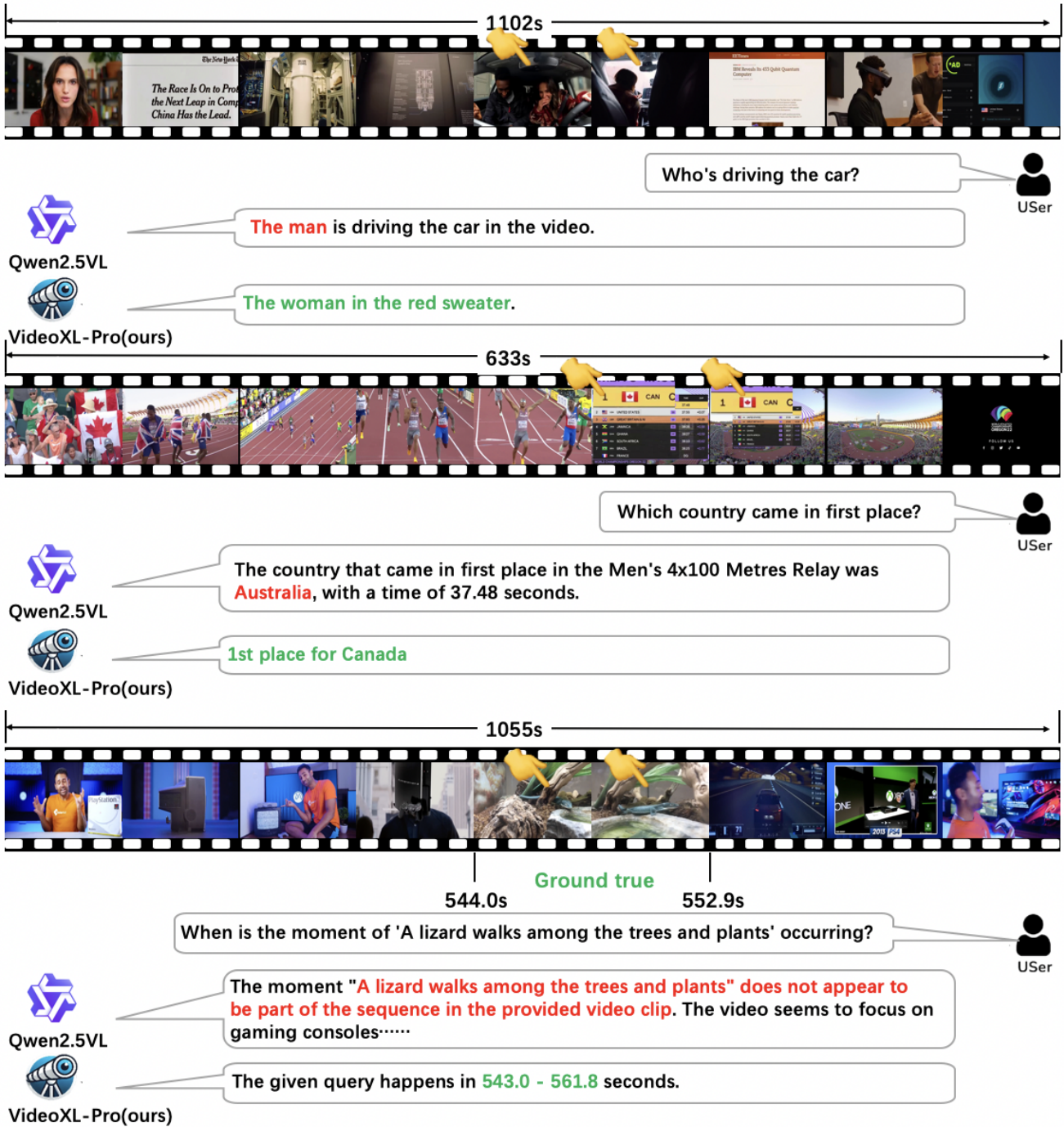}
    \caption{Quantitative evaluation of Video-XL-Pro in long video understanding.}
    \label{fig:vis}
\end{figure*}

\subsection{Temporal Grounding}
As detailed in Table~\ref{tab:time1} and Table~\ref{tab:time2}, compared to previous methods that rely on complex architectures to perceive timestamps, our method demonstrates surprisingly strong temporal grounding capabilities, especially on the long video temporal grounding benchmark V-STaR, where it almost outperforms the existing 7B MLLMs and achieves highly competitive performance.

\subsection{Extra-Long Evaluation}
To explore the capability of Video-XL-Pro in handling ultra-long video inputs, we further conducted a Needle-In-The-Haystack \cite{longva} evaluation based on the A100-80GB GPU. In the evaluations, MLLMs are required to trace along the chain of indicating images, locate the needle, and answer its corresponding question. We considered four model baselines in our assessment: 1) LLaVA-NexT \cite{zhang2024llavanextvideo} Video and LongLLaVA \cite{wang2024longllava}, which rely on positional extrapolation to extend to longer inputs; 2) LongVA \cite{longva}, which fine-tunes MLLM to handle longer inputs; and 3) VideoXL, which leverages the inherent key-value (KV) sparsification capability of MLLM to compress visual inputs. As illustrated in Figure \ref{fig:needle}, thanks to the strong capacity of ReCoT, Video-XL-Pro exhibits a disruptive advantage compared to the baselines. Due to computational cost constraints, neither LLaVA-NexT-Video nor LongLLaVA can support videos exceeding 1000 frames, while LongVA, after fine-tuning, can only support videos with fewer than 400 frames. Compared to the baseline Video-XL which can handle videos up to 2048 frames, Video-XL-Pro can cover inputs exceeding 8000 frames while maintaining an accuracy rate of nearly 99\%. This sufficiently demonstrates that our model achieves an astonishing balance between efficiency and effectiveness.



\subsection{Efficiency Analysis}
We further evaluate the training and inference efficiency of Video-XL-Pro. As illustrated in Figure \ref{fig:data_duration}, Video-XL-Pro significantly accelerates both training and inference by exponentially compressing visual information through ReCoT. Compared to the baseline model, Video-XL-Pro improves training speed by more than 2 times and enhances inference speed by nearly 3 times, while simultaneously achieving performance gains.


\subsection{Ablation Studies}
We conduct an extensive ablation study to evaluate the effectiveness of Video-XL-Pro, in particular the design of ReCoT, including Dynamic Token Synthesizer (DTS) and Semantic Guidance Masking (SGM). In addition, we demonstrate the effectiveness of the proposed pruning strategy for video datasets. 

\textbf{Dynamic Token Synthesize (DTS)}. To validate the effectiveness of DTS, we compare it with the strong baseline SigLIP. Specifically, we adopt a linear probing strategy, utilizing either SigLIP or our pretrained encoder (SigLIP + DTS) as a frozen feature extractor, followed by a trainable two-layer MLP for classification. We evaluate performance on three image benchmarks and three video benchmarks, as shown in Table \ref{tab:DTSn}. Results indicate that with the assistance of DTS, our pretrained visual encoder significantly outperforms SigLIP on multiple image benchmarks. Moreover, even when compressed by a factor of four, the DTS-enhanced encoder still achieves better results than SigLIP on video classification tasks. This demonstrates that our approach enables static image tokens to learn dynamic video features through spatio-temporal enhancement.




\textbf{Reconstructive Token Compression (ReCoT). }
To distinguish the contribution of ReCoT, we fine-tune Video-XL-Pro with and without ReCoT and evaluate its performance on two popular video understanding benchmarks, MLVU and VideoMME, as well as two image VQA benchmarks, MME \cite{fu2023mme} and MMB \cite{liu2023mmbench}. Due to computational resource constraints, we select a small dataset (50K images and 50K videos) for the ablation study. The empirical results, presented in Table \ref{tab:DTSVLM}, indicate that despite generating fewer visual tokens, the model with ReCoT still achieves better performance on both image and video benchmarks. These results demonstrate that ReCoT effectively enhances the representational capacity of SigLIP features, reduces semantic redundancy, and improves feature granularity, thereby enabling the model to perform long-term relational reasoning more effectively and better comprehend long videos. 

To demonstrate the contribution of Semantic-Guided Masking, we fixed the compression rate of DTS and compared it with a random mask at the same ratio. As shown in Table ~\ref{tab:SGM}, DTS with Semantic-Guided Masking achieved improvements on both image and video understanding benchmarks. This indicates that Semantic-Guided Masking can enhance the important features in visual understanding for DTS, thereby strengthening the model's comprehension ability.

\begin{table}[h]
    \centering
    \resizebox{1\linewidth}{!}{
    \renewcommand{\arraystretch}{1.15}
    \begin{tabular}{>{\kern-0.5\tabcolsep}ll|ccc<{\kern-0.5\tabcolsep}}
        \toprule
        \textbf{Model} & \textbf{Token} & \textbf{CIFAR10/100} & \textbf{Tiny-ImageNet(500)} & \textbf{Caltech 256}  \\
        \midrule
        SigLIP & 576 & 98.07/88.65 & 80.59 & 98.91  \\
        SigLIP+DTS ($2\times$) & 576 & 98.07/88.65 & 80.59 & 98.91  \\
        \rowcolor{ModelGreen} SigLIP+DTS ($4\times$) & 576 & \textbf{98.17/89.22} & \textbf{81.07} & \textbf{99.33}  \\
        SigLIP+DTS ($8\times$) & 576 & 98.07/88.65 & 80.59 & 98.91  \\
        \midrule
        \textbf{Model} & \textbf{Token} & \textbf{HMDB} & \textbf{K400(10\%)} & \textbf{K600(10\%)} \\
        \midrule
        SigLIP & 576 & 92.59 & 67.13 & 67.9 \\
        SigLIP+DTS ($2\times$) & 256 & 93.97 & 67.56 & 68.3 \\
        \rowcolor{ModelGreen} SigLIP+DTS ($4\times$) & \textbf{144} & \textbf{95.49} &\textbf{67.82} & \textbf{68.7} \\
        SigLIP+DTS ($8\times$) & 72 & 89.27 & 65.94 & 66.4 \\
        \midrule
        \bottomrule
    \end{tabular}
    }
    \caption{Effectiveness of Reconstructive Token Compression Method.}
    \label{tab:DTSn}
\end{table}

\begin{table}[h] 
 \centering
    \resizebox{1\linewidth}{!}{
    \renewcommand{\arraystretch}{1.15}
    \begin{tabular}{>{\kern-0.5\tabcolsep}ll|cccc<{\kern-0.5\tabcolsep}}
        \toprule
        \textbf{Model} & \textbf{Token} & \textbf{MLVU} & \textbf{VideoMME} & \textbf{MME} & \textbf{MMB} \\
        \midrule
        SigLIP & 576 & 50.4 & 45.8 & 1230.4 & 45.7 \\
        SigLIP+DTS ($2\times$) & 256 & 52.0 & 46.7 & 1256.7 & 46.7 \\
        \rowcolor{ModelGreen} SigLIP+DTS ($4\times$) & 144 & \textbf{53.1} & \textbf{47.5} & \textbf{1288.7} & \textbf{47.2} \\
        SigLIP+DTS ($8\times$) & 72 & 46.2 & 43.1 & 1224.9 & 45.3 \\
        \midrule
        \bottomrule
    \end{tabular}
    }
    \caption{Analysis of Reconstructive Token Compression in VLM. }\label{tab:DTSVLM}
\end{table}

\begin{table}[h] 
 \centering
    \resizebox{1\linewidth}{!}{
    \renewcommand{\arraystretch}{1.15}
    \begin{tabular}{>{\kern-0.5\tabcolsep}l|cccc<{\kern-0.5\tabcolsep}}
        \toprule
        \textbf{Model} & \textbf{MLVU} & \textbf{VideoMME} & \textbf{MME} & \textbf{MMB} \\
        \midrule
        SigLIP+DTS+RandomMask & 52.4 & 47.0 & 1260.4 & 46.1 \\
        \rowcolor{ModelGreen} SigLIP+DTS+SGM & \textbf{53.1} & \textbf{47.5} & \textbf{1288.7} & \textbf{47.2} \\
        \midrule
        \bottomrule
    \end{tabular}
    }
    \caption{Analysis of Semantic-Guided Masking in ReCot. }\label{tab:SGM}
\end{table}

\textbf{Video Dataset Pruning}
To distinguish the contribution of Video Dataset Pruning, we constructed three SFT datasets to fine-tune Video-XL-Pro. As shown in the Table~\ref{tab:videopurning}, the 100k SFT data processed by Video Dataset Pruning achieves comparable results to randomly selected 200k data, particularly on MLVU. This indicates that our Video Dataset Pruning can significantly improve data quality, as multi-scenario data enables the model to learn more knowledge. Video Dataset Pruning allows Video-XL-Pro to achieve state-of-the-art performance with a relatively small training scale.

\begin{table}[h]
 \centering
    \resizebox{1\linewidth}{!}{
    \renewcommand{\arraystretch}{1.15}
    \begin{tabular}{>{\kern-0.5\tabcolsep}ll|ccc<{\kern-0.5\tabcolsep}}
        \toprule
        \textbf{Model}& \textbf{Training-time} & \textbf{VideoMME} & \textbf{MLVU} & \textbf{LongVideoBench} \\
        \midrule
        Random100k & 40 GPUhours & 45.4 & 49.2 & 42.1 \\
        Random200k & 78 GPUhours & 47.9 & \textbf{52.0} & 44.3 \\
        \rowcolor{ModelGreen} Pruning100k & 40 GPUhours & \textbf{48.1}  & 51.5 & \textbf{44.6} \\
        \midrule
        \bottomrule
    \end{tabular}
    }
    \caption{Analysis on Video Dataset Pruning.}\label{tab:videopurning}
\end{table}

\begin{figure}[h]
    \centering
    \includegraphics[width=0.5\textwidth]{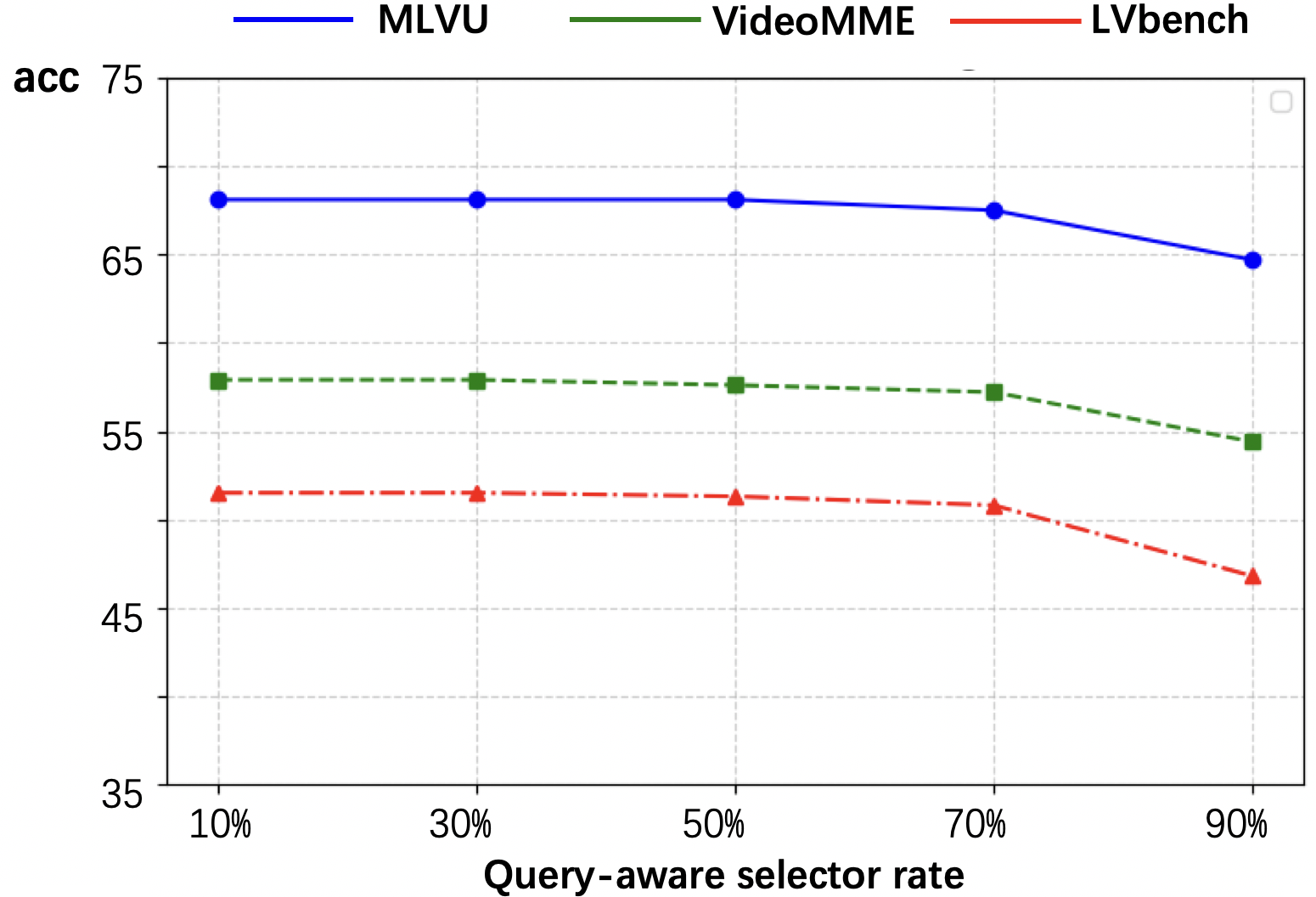}
    \caption{Different Query-aware selector rates. ``LVbench" refers to LongVideoBench.}
    \label{fig:query}
\end{figure}

\textbf{Query-aware selector rate.} Finally, to evaluate the contribution of the Query-Aware Selector rate to model performance, we conduct experiments using different Query-Aware Selector rates and assess them on three popular long-video understanding benchmarks: MLVU, VideoMME, and LongVideoBench. As shown in Figure~\ref{fig:query}, the empirical results indicate that the Query-Aware Selector can compress tokens to 50\% of their original size without any information loss. Even at an extremely high compression rate (90\%), it still maintains considerable accuracy. These findings suggest that the Query-Aware Selector effectively identifies query-relevant tokens while reducing redundant information.


\subsection{Qualitative Results}
We employ qualitative evaluation to intuitively analyze Video-XL-Pro. In this experiment, we compare it with Qwen2.5-VL-7B based on ultra-long videos (with durations exceeding 30 minutes). As shown in Figure~\ref{fig:vis}, Video-XL-Pro is capable of accurately locating key information in long videos and presenting its details; in contrast, Qwen2.5-VL struggles with understanding the content of long videos and fine-grained details. Video-XL-Pro effectively locates the required key information in long videos. This further validates the robustness of Video-XL-Pro in long video understanding tasks.

\section{Conclusion}
In this work, we introduced Video-XL-Pro, a novel and efficient method designed to address the challenges of long video understanding. By proposing Reconstructive Compression of Tokens (ReCoT), a module that leverages self-supervised learning, we significantly enhanced the quality and efficiency of video comprehension. The two key components of ReCoT—Dynamic Token Synthesizer (DTS) and Semantic-Guided Masking (SGM)—enable the generation of comprehensive and compact video tokens, improving both the quality and efficiency of video understanding. Additionally, our video-specific dataset pruning strategy and post-processing mechanism ensure higher-quality training data and precise localization of query-relevant video tokens, respectively. Despite its compact size of only 3B parameters, Video-XL-Pro outperforms larger models on multiple long video understanding benchmarks, demonstrating its effectiveness.


\clearpage

{
    \small
    \bibliographystyle{ieeenat_fullname}
    \bibliography{main}
}


\clearpage

\setcounter{page}{1}
\maketitlesupplementary

\appendix 
\setcounter{figure}{0} 
\setcounter{table}{0}

\section*{Overview of Supplementary Material}

\begin{itemize}

    \item  \ref{appendix:Training Datasets}: \textbf{Training Datasets}
    \item \ref{appendix:experiment}: \textbf{Experimental Settings and Additional Results}

\end{itemize}

\section{Training Datasets}
 \label{appendix:Training Datasets} 

\textbf{Video-XL-Pro Training Datasets}

Video-XL-Pro employs a two-stage training approach. In the first stage, we freeze SigLIP, DTS, and LLM, and only train the projection module. In the second stage, we perform full-parameter fine-tuning. During both stages of training, we leverage large-scale video-text pairs from multiple publicly accessible databases,including LLaVA-OneVision~\cite{li2024llavaonevision},  LAION-2M~\cite{he2024bunny}, Charades~\cite{sigurdsson2016charades}, CinePine~\cite{rawal2024cinepile}, OpenViD~\cite{nan2025openvid1m}  ShareGPT4o~\cite{sharegpt4o}, Ego4d~\cite{mangalam2023egoschema}, Ocrvqa~\cite{Anand2019ocrvqa}, Next-QA~\cite{xiao2021nextqanextphasequestionansweringexplaining},  LLaVA178k~\cite{zhang2024videoinstructiontuningsynthetic} and all video datasets are compressed using our Video Dataset Pruning method to ensure extremely high-quality training data. All training data is detailed in Table~\ref{tab:VideoXLProTraining}.


\begin{table}[h] 
 \centering
    \vspace{-0.1in}
    \resizebox{1\linewidth}{!}{
    \renewcommand{\arraystretch}{1.15}
    \begin{tabular}{c|c|cc}
    \toprule
    \textbf{Modality} & \multicolumn{1}{c|}{\textbf{Type}} & \textbf{\# Samples} & \textbf{Dataset} \\ 
    \hline
     \multirow{4}{*}{Pretrain} & \cellcolor{gray!10} Image     & \cellcolor{gray!10} 0.8M & \cellcolor{gray!10} LLaVA-OneVision(sub) \\
      & Image     & 0.8M  & LAION-2M(sub)  \\
      & \cellcolor{gray!10} Video     & \cellcolor{gray!10} 10k & \cellcolor{gray!10} Charades \\
      & Video  & 300k  & OpenViD(sub)  \\

    \hline
     & Image     & 21K  & GPT-4o-image \\
     & \cellcolor{gray!10} Video & 2K \cellcolor{gray!10}  & \cellcolor{gray!10} GPT-4o-video \\
     & Video     & 28K  & CinePine \\
     & \cellcolor{gray!10} Video & 38K \cellcolor{gray!10}  & \cellcolor{gray!10} Sharegpt4v \\
     & Video     & 0.7K  & Ego-4d \\
     & \cellcolor{gray!10} Video & 43K \cellcolor{gray!10}  & \cellcolor{gray!10} FineVideo \\
     & Image     & 19K  & OCR-VQA \\
     & \cellcolor{gray!10} Video & 3K \cellcolor{gray!10}  & \cellcolor{gray!10} ScienceQA \\
     & Video    & 107K  & VCG \\
     & \cellcolor{gray!10} Video & 15K \cellcolor{gray!10}  & \cellcolor{gray!10} ActivitynetQA \\
     & Image   & 390K  & Bunny695k \\
     & \cellcolor{gray!10} Video & 34K \cellcolor{gray!10}  & \cellcolor{gray!10} Next-QA \\
     & Video   & 281K  & LLaVA178k \\
     
    \multirow{-15}{*}{SFT} & \cellcolor{gray!10} Video & \cellcolor{gray!10}  1K & \cellcolor{gray!10}  \begin{tabular}[c]{@{}l@{}} VDC \end{tabular} \\ 
    \bottomrule
    \end{tabular}
    }
    \vspace{-0.1in}
    \caption{All Training Datasets }\label{tab:VideoXLProTraining}
\end{table}

\textbf{Reconstructive Compression of Tokens (ReCoT) Training Datasets}

We employ self-supervised learning to conduct Reconstructive Compression of Tokens (ReCoT) training, guided efficiently by Semantic-Guided Masking. Leveraging data from multiple publicly accessible datasets, including Bunny, llava178k and OCRVQA, we successfully complete ReCoT training using only 1 million data points. This approach achieves superior performance compared to the original SigLIP features across multiple image and video benchmarks.All training data is detailed in Table~\ref{tab:ReCotTraining}.
\begin{table}[h] 
 \centering
    \vspace{0.1in}
    \resizebox{1\linewidth}{!}{
    \renewcommand{\arraystretch}{1.15}
    \begin{tabular}{c|c|cc}
    \toprule
    \textbf{Modality} & \multicolumn{1}{c|}{\textbf{Type}} & \textbf{\# Samples} & \textbf{Dataset} \\ 
    \hline
     \multirow{2}{*}{ReCot train} & \cellcolor{gray!10} Image     & \cellcolor{gray!10} 250k & \cellcolor{gray!10} Bunny695k,OCR-VQA \\
      & Video     & 750k  & LLaVA178k  \\

    \bottomrule
    \end{tabular}
    }
    \vspace{-0.1in}
    \caption{ReCot Training Datasets }\label{tab:ReCotTraining}
\end{table}

\section{Experimental Settings \& Additional Results}
\label{appendix:experiment}
We elaborate on the training and inference details of Video-XL-Pro. Since our method only modifies the workflow of LLM, the hyperparameters reported are specific to the fine-tuning stage, as shown in Table~\ref{tab:hyper}. 

   


\begin{table}[h]
\centering
    \centering
   
    \vspace{-0.1in}
    \renewcommand{\arraystretch}{1.15}

    \begin{tabular}{>{\kern-0.5\tabcolsep}l|c<{\kern-0.5\tabcolsep}}
        \toprule
        \textbf{Hyperparameter}  & \textbf{Value}   \\
        \midrule
         Overall batch size  &64  \\
        Learning rate  & 1e-5 \\
        LR Scheduler  & Cosine decay \\
        DeepSpeed ZeRO Stage &ZeRO-2-offload \\
        Optimizer & Adam \\
        Warmup ratio & 0.3 \\
        Epoch & 1 \\
        Weight decay & 0 \\
        Precision & bf16 \\ 
        \bottomrule
    \end{tabular}
     \caption{Hyperparameters of Video-XL-Pro}
    \label{tab:hyper}
\end{table}

    


{
    \small
    \bibliographystyle{ieeenat_fullname}
    \bibliography{main}
}

\end{document}